\documentclass[10pt,twocolumn,letterpaper]{article}

\usepackage{cvpr}              
\usepackage{balance}

\usepackage[table]{xcolor}
\usepackage{multirow}
\usepackage{arydshln}   

\definecolor{cvprblue}{rgb}{0.21,0.49,0.74}
\usepackage[pagebackref,breaklinks,colorlinks,allcolors=cvprblue]{hyperref}

\usepackage{graphicx}
\usepackage[ruled,vlined,noend]{algorithm2e}
\usepackage{multirow} 
\usepackage{cuted} 
\usepackage{array}

\def\paperID{} 
\def\confName{CVPR}
\def\confYear{2026}

\title{Frame2Freq: Spectral Adapters for Fine-Grained Video Understanding}
\author{
Thinesh Thiyakesan Ponbagavathi\textsuperscript{1,3} \quad
Constantin Seibold\textsuperscript{2} \quad
Alina Roitberg\textsuperscript{3} \\
\textsuperscript{1}\small Institute for Artificial Intelligence, University of Stuttgart, Germany \\
\textsuperscript{2}\small University Hospital Heidelberg, Diagnostic and Interventional Radiology \\
\textsuperscript{3}\small Intelligent Assistive Systems Lab, University of Hildesheim
}

\begin{document}
\maketitle
\begin{abstract}

Adapting image-pretrained backbones to video typically relies on time-domain adapters tuned to a single temporal scale. Our experiments show that these modules pick up static image cues and very fast flicker changes, while overlooking medium-speed motion. Capturing dynamics across multiple time-scales is, however, crucial for \textsl{fine-grained} temporal analysis (i.e., \textsl{opening} vs. \textsl{closing bottle}).

To address this, we introduce Frame2Freq -- a family of frequency-aware adapters that perform spectral encoding during image-to-video adaptation of pretrained Vision Foundation Models (VFMs), improving fine-grained action recognition. Frame2Freq uses Fast Fourier Transform (FFT) along time and learns frequency-band specific embeddings that adaptively highlight the most discriminative frequency ranges.
 Across five fine-grained activity recognition datasets, Frame2Freq  outperforms prior PEFT methods and even surpasses fully fine-tuned models on four of them. These results provide encouraging evidence that frequency analysis methods are a powerful tool for modeling temporal dynamics in image-to-video transfer. Code is available at https://github.com/th-nesh/Frame2Freq.

\end{abstract}


\section{Introduction}
\label{sec:intro}
\begin{figure}[t!]
    \centering
\includegraphics[width=1\columnwidth]{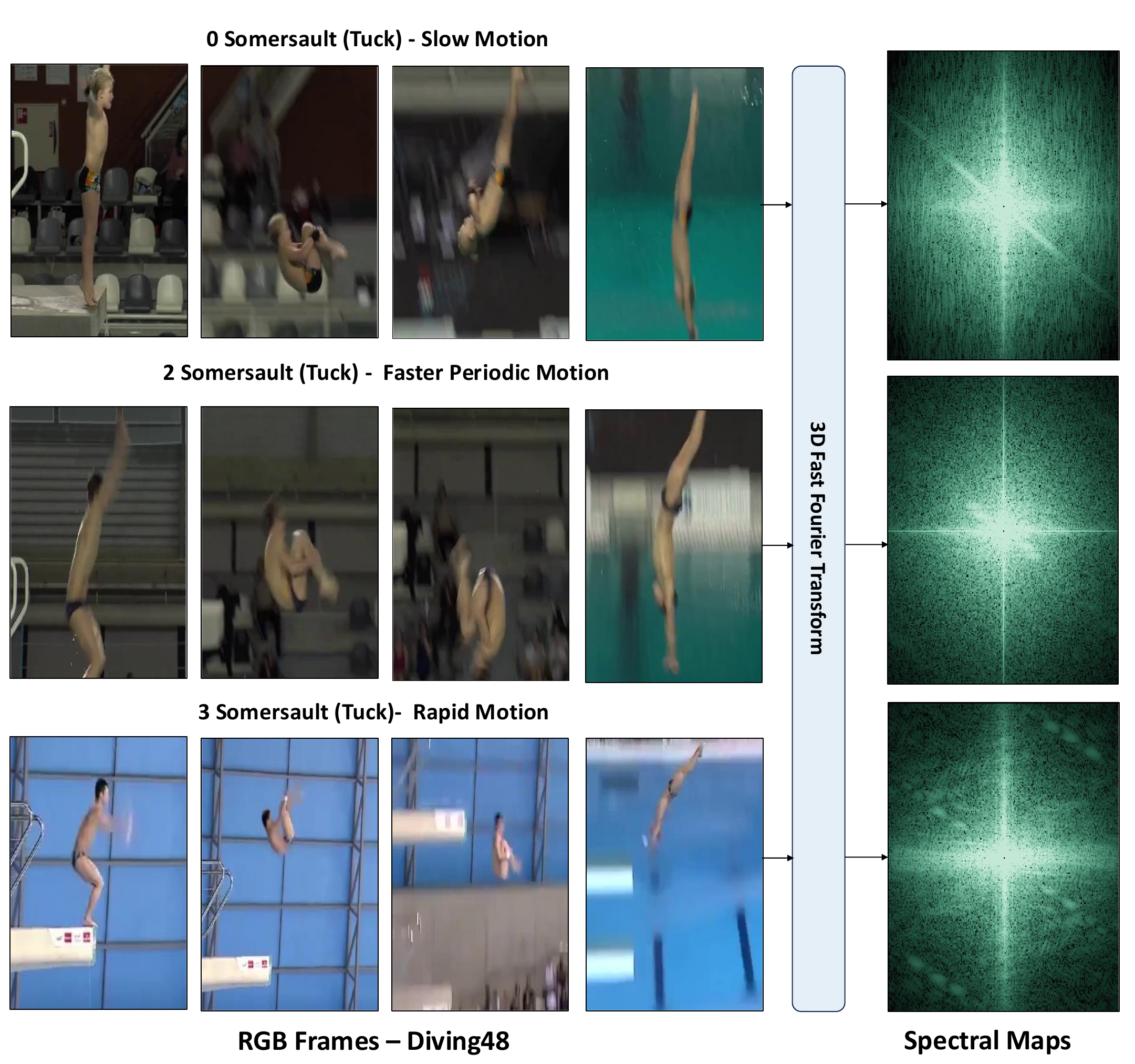}
\caption{
Examples from Diving48 illustrating how different somersault counts induce distinct temporal frequency signatures. 
Spectral magnitude maps on the right are obtained by applying a  Fast Fourier Transform temporally to the frame embeddings of a frozen VFM.   Slow dives produce low-frequency spectra, while faster periodic and rapid motions shift energy toward mid and high frequencies. 
This  motivates Frame2Freq -- our  image-to-video PEFT approach, leveraging Fast Fourier Transform
(FFT) along time to improve fine-grained video understanding.
}
\label{fig:teaser}
\vspace{-1em}
\end{figure}


Existing parameter-efficient image-to-video adapters address temporal reasoning through time-axis convolutions \cite{St_adaptor,dualpath} or attention fusion across frames~\cite{omniclip}. 
We take a different route and introduce \textit{ frequency-aware adapters} that model motion in the spectral domain, enabling pretrained Vision Foundation Models (VFMs) \cite{bommasani2021opportunities} to better capture the rhythm and scale of temporal change.
First, we validate the need for finer temporal analysis by inspecting the frequency content of CLIP embeddings under common image-to-video adapters and show that they concentrate energy at the low- and high-frequency bands and overlook the mid-frequencies.
This limits \textit{fine-grained }video understanding (e.g., sports analysis~\cite{finegym}), where subtle phase shifts 
during motion  define the discriminative signal. Such fine actions exhibit characteristic frequency patterns (Figure~\ref{fig:teaser}), making the spectral domain useful for temporal adaptation. 

Understanding fine-grained human actions such as different variants of a \textit{somersault} in Figure (\ref{fig:teaser})  requires sensitivity to subtle motion intent and contact dynamics \cite{tqn,finegym}. Even harder are nearly symmetric actions \cite{STEP,chirality}, like \textit{pick up an object} vs. \textit{lay down an object}, which share almost identical spatial configurations but differ in motion phase. These distinctions matter for several applications, ranging from sports analysis \cite{diving48} to driver monitoring \cite{drive_and_act} and collaborative robotics \cite{IKEA_ASM,hri_30}, where systems must track motion dynamics and task progression. Recent advances VFMs, such as CLIP \cite{CLIP} and DINOv2 \cite{dinov2}, have shown impressive generalization from images to videos, inspiring numerous adaptations \cite{St_adaptor,aim,VitaCLIP,m2_clip,omniclip,EVL,ILA} for temporal reasoning. Yet, they still struggle to encode the fine temporal cues that define human motion.



Thus, we seek a novel spectral transforms-grounded adapter and introduce Frame2Freq, a family of frequency-aware adapters for image-to-video transfer.  
Temporal embeddings from frozen VFM backbone are transformed via a Fast Fourier transform (FFT) to extract frequency patterns and then reconstructed into phase-aligned representations that capture intricate motion dynamics. Two variants: Frame2Freq-ST and Frame2Freq-MS, extend this design across single and multi-scale temporal windows, adapting to datasets with distinct motion characteristics. Our ANOVA \cite{annova} inspired Frequency Discriminability analysis shows that Frame2Freq allocates representational energy toward mid-frequency bands, aligning model attention to spectral regions most correlated with fine-grained actions.



Our contributions are as following:

(1) To the best of our knowledge, we for the first time explore spectral transforms and frequency analysis as a basis for image-to-video transfer in pretrained Vision Foundation Models (VFMs). To motivate this direction, we conduct a Frequency Discriminability Analysis that quantifies per-band discriminative energy and reveals that mid-range frequencies are most informative for fine-grained motion understanding.
(2) Building on this insight, we propose Frame2Freq, the first family of frequency-domain adapters that adapt image-pretrained VFMs to video tasks, improving fine temporal reasoning without retraining spatial weights.
(3) Extensive experiments across five fine-grained activity recognition benchmarks (SSv2 and Diving48, Drive\&Act, IKEA-ASM, and HRI30) demonstrate consistent gains over fully fine-tuned and PEFT baselines while requiring minimal trainable parameters.  Together, these results highlight that frequency structure is key to bridging the gap between static vision models and dynamic video understanding.


\section{Related Works}
\label{sec:related_works}
 \noindent\textbf{Fourier Transform in Computer Vision.}
Fourier and other spectral transforms have a long history in vision research.
Early CNN works exploited FFT to accelerate convolutions \cite{mathieu2014fasttrainingconvolutionalnetworks} and  design information-preserving pooling \cite{spectral_pooling}, laying the groundwork for spectral reasoning in deep networks. Fast Fourier Convolution \cite{Fast_fourier_Convolution} combined spatial and frequency operators, merging local convolution with global receptive fields. Subsequent studies \cite{frequency_cnn,frequency_pruning,Frequency_prune_2,Frequency_prune_3} leveraged the low-frequency bias of CNNs for spectral pruning to reduce redundancy, focusing on compression rather than richer representations.


With the shift to transformers, GFNet \cite{GFNet} and Spectformer \cite{Spectformer}  replaced attention with learnable Fourier filters, achieving competitive accuracy with logarithmic complexity via FFT/IFFT token mixing. Dynamic Temporal Filtering \cite{DTF}  extended this idea to the temporal axis through 1D FFT-based filters, improving video models but requiring full model fine-tuning,  unsuitable for adapting VFMs.
More recent approaches, including Visual Fourier Prompt Tuning \cite{VFPT}, FADA \cite{fada}, and Fourier-VLM \cite{fourier_vlm}, use frequency information for tasks like domain adaptation or token compression. Among them, VFPT is the only parameter-efficient tuning (PEFT) approach that explicitly uses frequency cues, but only for spatial adaptation. No prior work explores frequency structure to guide temporal reasoning in frozen image backbones. Our work bridges this gap by introducing multi-scale frequency transforms within a PEFT framework, aligning frozen VFMs with the temporal frequency structure of real-world actions.


\noindent\textbf{Adapting Vision Foundation Models for Image-to-Video Transfer.} Downstream task adaptation of VFMs follows two paradigms: \textit{Probing} and \textit{Parameter Efficient Fine-tuning (PEFT)}.
Probing learns lightweight heads such as linear \cite{linear_prob_simclr, linear_prob_selfsupervisedvisualrepresentation} or attention probes \cite{v-jepa,set_transformer} on top of frozen pretrained backbones. STEP \cite{STEP} extends this idea with temporally-aware probes that improve sensitivity on smaller, domain-specific datasets but lacks capacity for large-scale generalization.

In contrast, PEFT introduces few learnable parameters that guide the VFM in downstream tasks, while keeping the backbone frozen. Adapter-based approaches such as ST-Adapter \cite{St_adaptor}, AIM \cite{aim}, M2-CLIP \cite{m2_clip}, and Omni-CLIP \cite{omniclip} learn temporal dependencies using convolutional layers, non-linear activation, temporal difference blocks, and attention blocks. Fully-finetuned methods like STAN \cite{STAN}, BIKE \cite{BIKE}, and ILA \cite{ILA} improve temporal reasoning by adding auxiliary networks, bidirectional cross-modal alignment, and temporal alignment via learnable masks. Yet all treat motion as sequential frame differences, ignoring its frequency structure and thereby limiting generalization on fine-grained video understanding where discriminative cues reside in specific frequency bands. Our work falls into the PEFT category and serves as an image-to-video adaptation method. 
To the best of our knowledge, Frame2Freq is the first PEFT-based image-to-video adapter that models temporal motion  in the frequency domain.


\section{Methodology}
\label{sec:Methods}

\begin{figure*}
    \centering
\includegraphics[width=1\linewidth]{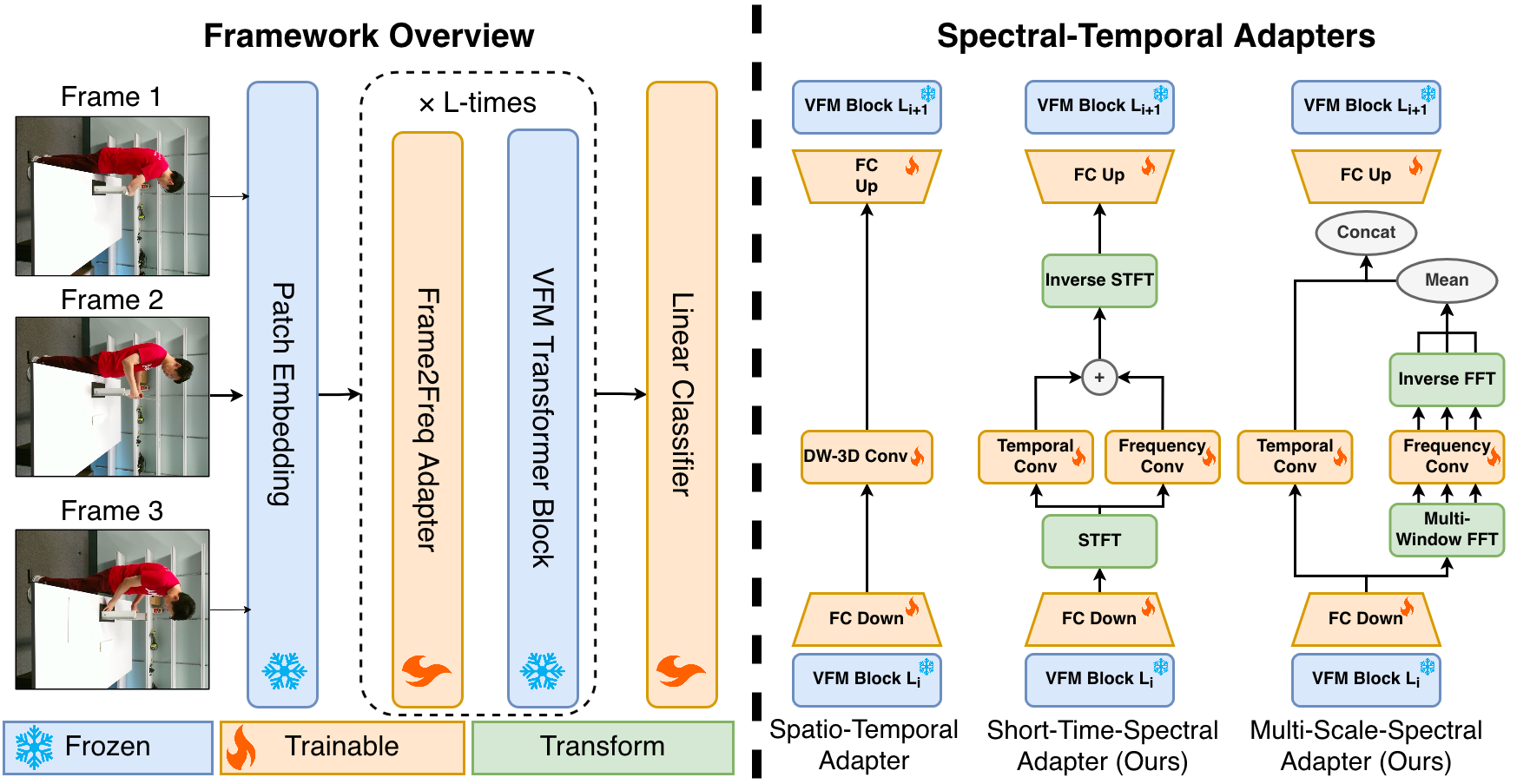}
\caption{\textbf{Overview of Frame2Freq.} Left: Frozen VFMs are adapted to video by inserting lightweight Frame2Freq adapters between transformer blocks, enriching spatial embeddings with frequency-aware temporal cues. Right: Unlike the ST-Adapter, our spectral variants use FFT-based branches: Frame2Freq-ST for single scale spectral clues and Frame2Freq-MS for multi-scale motion patterns. 
}
\label{fig:architecture}
\end{figure*}

Our idea is to explicitly model the \textit{frequency} structure of the temporal dimension when adapting frozen VFMs for video recognition. We introduce Frame2Freq, a family of frequency-aware adapters that are built within a PEFT framework and inject spectral reasoning into frozen backbones for finer temporal understanding. By incorporating different frequency transforms, we derive two variants: Frame2Freq-ST and Frame2Freq-MS. 

\subsection{Framework Overview}
Adapting Image VFMs for video understanding requires adding temporal sensitivity without retraining millions of parameters. Frame2Freq achieves this by modulating VFM embeddings through frequency-aware cues. Each input video is represented as a sequence of frames $
V = \{I_1, I_2, \dots, I_T\}, \quad I_t \in \mathbb{R}^{H \times W \times 3},$
which are tokenized by a frozen Vision Transformer backbone (e.g., CLIP \cite{CLIP} or DINOv2 \cite{dinov2}) into patch embeddings
$
X = \{x_t^n\} \in \mathbb{R}^{T \times N \times D},
$
where $T$ is the number of frames, $N$ the spatial tokens per frame, and $D$ the embedding dimension.
Each adapter follows a compact bottleneck structure ($\text{FC}_{down}$ $\rightarrow$ non-linearity $\rightarrow$ $\text{FC}_{up}$). Unlike ST-Adapter \cite{St_adaptor} -our main PEFT baseline shown on the left side of Figure \ref{fig:architecture}, which models motion using depthwise temporal convolutions in the time domain,
Frame2Freq introduces a spectral path that transforms temporal features into the frequency domain, learns band-specific filters to highlight mid-frequency motion cues.
The resulting spectral features are projected to the original embedding dimension $D$ and added back to the backbone output through a residual connection, enriching each transformer block with frequency-aware temporal refinement.  Frame-wise CLS aggregation feeds a lightweight linear head for classification. This design preserves strong spatial priors while modeling discriminative temporal dynamics. Figure~\ref{fig:architecture} summarizes the framework.

\subsection{\textbf{Frame2Freq-ST}: Short-Time Spectral Adapter}
Frame2Freq-ST embeds localized spectral reasoning into the adapter pathway using the Short-Time Fourier Transform (STFT). The video embeddings $X$ are projected through $\text{FC}_{down}$, decomposed into temporal windows and converted into frequency space using STFT.
\begin{equation}
\tilde{X}(f, \tau) = \sum_{t=0}^{T-1} \text{FC}_{down}(X(t))\, w(t - \tau)\, e^{-j 2 \pi f t}
\end{equation}
where $\tilde{X} \in \mathbb{C}^{B \times N \times F \times T' \times C_a}$, \(w(t - \tau)\) is a Hann window centered at \(\tau\), \(f\) the frequency index, and \(C_a\)  the adapter channel dimension after the down-projection. The spectral tensor ($\tilde{X}$) encodes both temporal locality (via $\tau$) and frequency composition (via $f$). 
Two depthwise 3D convolutions then refine it: $\text{Conv}_{temp}$ captures short-term transitions, while $\text{Conv}_{freq}$ models relationships between neighboring bands. 
The refined representation is then projected back to the temporal domain using an inverse STFT (iSTFT) and then restored by $\text{FC}_{up}$. 
\begin{equation}
\hat{X}(t) = \text{FC}_{up} \left( \text{iSTFT}\big(\text{Conv}_{freq}(\tilde{X}) + \text{Conv}_{temp}(\tilde{X})\big) \right)
\end{equation}
The output $\hat{X}(t)$ is fused with the frozen backbone features through a residual connection to preserve spatial semantics. 
Fine-grained actions, such as hand-object interactions, exhibit motion concentrated in mid-frequency bands, while static or abrupt movements occupy low or high extremes. 
Frame2Freq-ST amplifies mid-frequency energy, suppressing irrelevant components to isolate subtle, discriminative motion patterns that distinguish fine-grained actions.

\subsection{\textbf{Frame2Freq-MS}: Multi-Scale Spectral Adapter}
While Frame2Freq-ST focuses on localized single scale spectral reasoning, Frame2Freq-MS extends it across multiple temporal resolutions to capture both fine and coarse motion.
Instead of a fixed window, as in STFT, it jointly models time and frequency by using two coordinated branches: one processes temporal features, while the other applies Fourier transforms at multiple window sizes to form a multi-scale spectral representation.
After $\text{FC}_{down}$, embedding channels $C_a$ are split into $X_{freq}$ and $X_{temp}$: $X_{freq}, X_{temp} = \text{Split}(\text{FC}_{down}(X))$. The frequency branch applies $K$ Fourier transforms using window sizes ${w_k}$:
\begin{equation}
\tilde{X}_k(f) = \sum_{t=0}^{T-1} X_{freq}(t) e^{-j 2 \pi f t / w_k}, \quad k = 1, \dots, K.
\end{equation}
Each  $\tilde{X}_k \in \mathbb{C}^{B \times N \times F_k \times C_a/2}$ is refined by a shared depthwise $(3\times1\times1)$ convolution $\text{Conv}_{freq}$ to enhance scale-consistent frequency responses. The filtered outputs are averaged and projected back via an inverse FFT:

\begin{equation}
\tilde{X}_{freq} = \text{iFFT}\left(\frac{1}{K}\sum_{k=1}^{K}\text{Conv}_{freq}(\tilde{X}_k)\right).
\end{equation}
In parallel, the temporal branch processes $X_{temp}$ using a matching $(3\times1\times1)$ convolution $\text{Conv}_{temp}$ that captures short-range temporal continuity.
\begin{equation}
\tilde{X}_{temp} = \text{Conv}_{temp}(X_{temp}), {X}_{temp} \in \mathbb{C}^{B \times T \times N \times C_a/2} 
\end{equation}

Finally, we concatenate the outputs of both branches to bring back the channel dimension to $FC_{down}$ and restore it to the original VFM embedding dimension through $FC_{up}$.

\begin{equation}
\hat{X}(t) = \text{FC}_{up}\big(\text{Concat}[\tilde{X}_{freq}, \tilde{X}_{temp}]\big).
\end{equation}
By unifying multi-scale spectral analysis with localized temporal refinement, Frame2Freq-MS balances fast and slow motion modeling, generalizing to complex datasets where action frequencies vary widely across contexts.

\section{Frequency Spectral Analysis}
\label{sec:spectral_analysis}

\begin{figure}[ht!]
    \centering
\includegraphics[width=1\columnwidth]{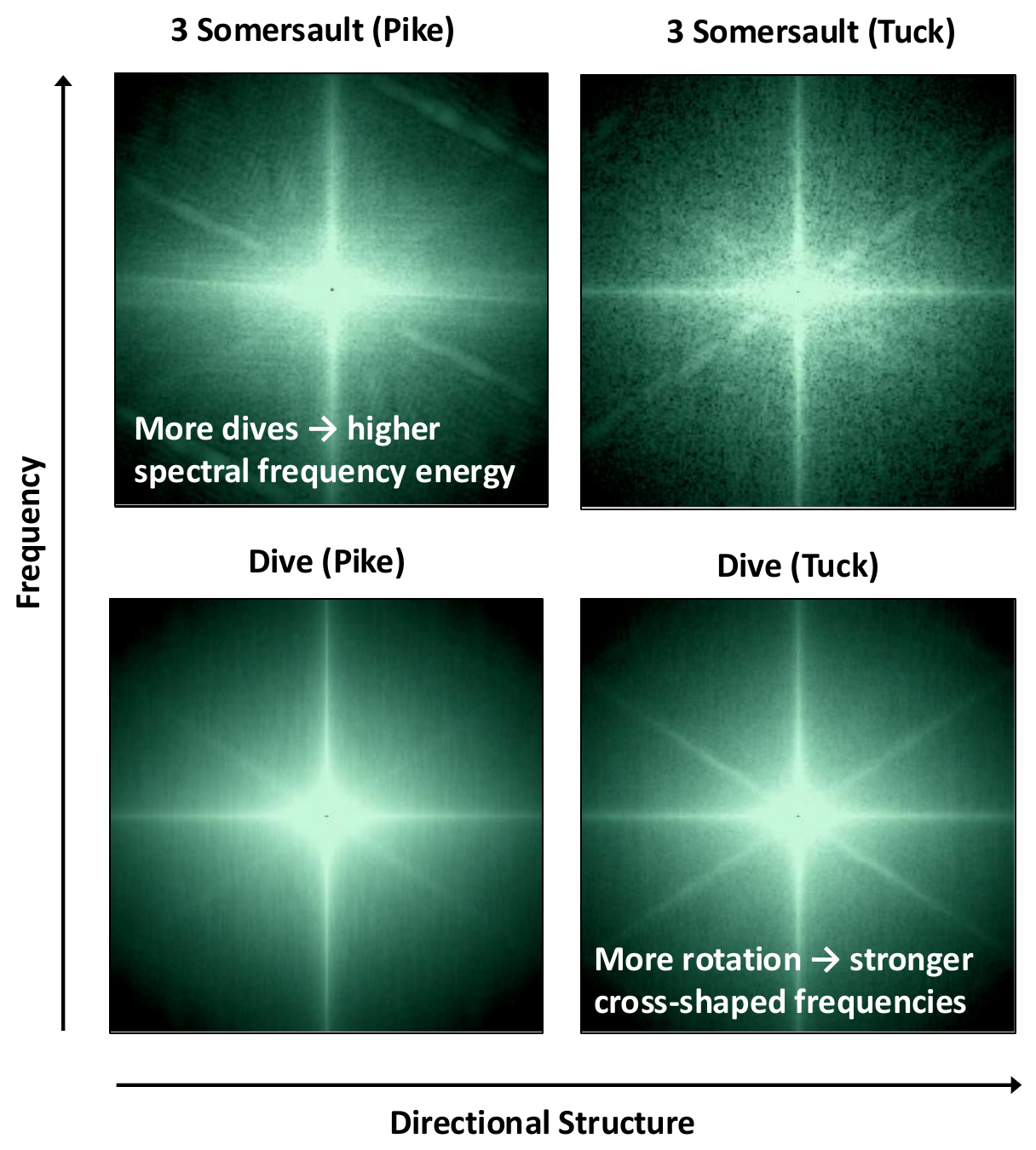}
\caption{Class-wise mean spectra from Diving48~\cite{diving48} reveal that dive complexity boosts frequency energy, while posture (tuck vs. pike) shapes directional components. Fine-grained distinctions emerge more cleanly in the frequency domain, motivating frequency-aware temporal modeling.
\vspace{-1.5em}
}

\label{fig:spectrum_analysis}
\end{figure}

Fine-grained actions hinge on subtle motion dynamics that are hard to capture in RGB space, motivating us to first examine video behavior in the frequency domain.
We use 3D Fast Fourier Transform on the videos and compute class-wise mean spectra on Diving48 \cite{diving48}. 
The results in Figure  \ref{fig:spectrum_analysis} confirm that actions with more motion cycles (\textit{e.g}., multiple somersaults) produce high spectral energy, while posture variations (tuck vs. pike) modulate the energy in the cross-shaped frequencies. These consistent, class-specific spectral patterns reveal motion cues not visible in the RGB-space and motivate using frequency representations for temporal adaptation of VFMs.

\begin{figure*}[t!]
    \centering
\includegraphics[width=1\textwidth]{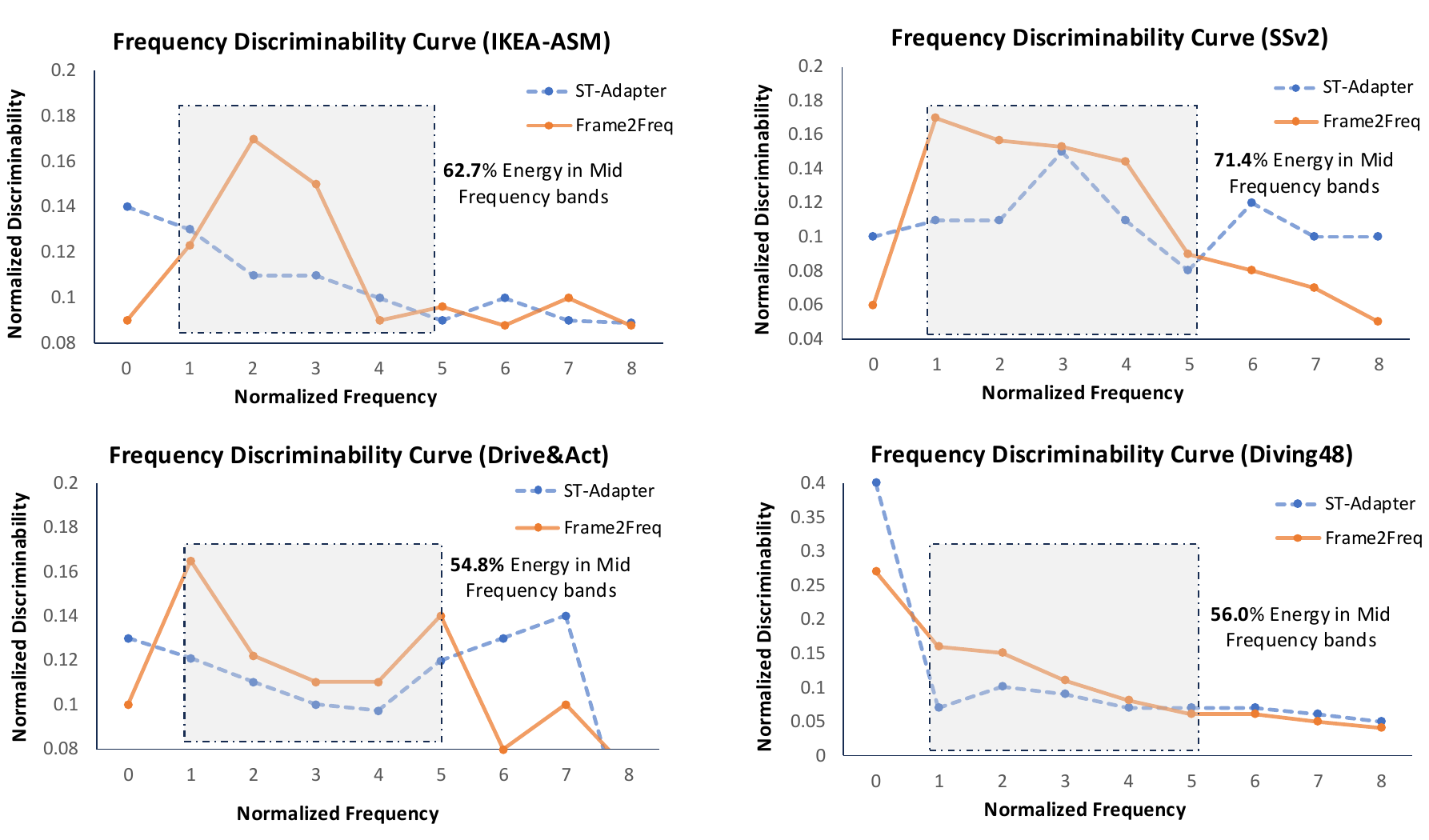}
\caption{
Normalized frequency discriminability curves $D(f)$ for our main baseline ST-Adapter~\cite{St_adaptor} (blue) and our Frame2Freq-adapter (orange) on four datasets, computed with the Frequency Discriminability Analysis described in Sec.~\ref{sec:spectral_analysis}. 
Each curve shows how much class-separating power is carried by each temporal frequency band (0-8). 
Standard temporal adapters concentrate discriminability in low or very high frequencies and tend to underuse the mid bands, whereas Frame2Freq  shifts discriminability toward the most informative bands for each dataset, which is especially useful for recognition of fine-grained actions.
}
\vspace{-1em}
\label{fig:freq_analysis}
\end{figure*}

To quantify this observation, we perform a Frequency Discriminability Analysis (detailed in the supplementary material), inspired by the Analysis of Variance (ANOVA) framework \cite{annova, anova_cause}. By decomposing variance across frequencies, we compute a discriminability score $D(f)$ that quantifies how well action classes separate at each temporal band. This provides a principled way to study the frequency structure underlying the representations learned by VFM adapters. Across datasets, temporal adapters exhibit a strong spectral bias:  they concentrate energy at very low or high frequencies, while underutilizing the mid-frequency bands where fine-grained dynamics typically reside (Figure \ref{fig:freq_analysis}).
In our setting (16 frames at 30fps), we define mid-frequency components as FFT bins 1–5 ($\approx$1--10Hz), a range known to contribute strongly to motion sensitivity \cite{Watson1985-cn}. These mid-band regions capture structured motion cues that distinguish actions such as \textsl{opening} vs.\ \textsl{closing} or \textsl{tuck} vs.\ \textsl{pike}, scenarios where existing temporal adapters struggle.

Frame2Freq corrects this imbalance. Rather than imposing a fixed spectral profile, it adapts its discriminability curve to the motion characteristics of each dataset.
In Figure \ref{fig:freq_analysis}, the orange curves show different shapes across all four datasets: SSv2 concentrates most of the discriminative power in the upper mid-frequency bands, reflecting its fast object interactions. IKEA-ASM and Drive\&Act peak in the mid-range frequencies associated with human–object manipulation.  Diving48 shifts its emphasis toward the lower end of the mid band, consistent with the dataset’s small frame-to-frame differences and slow, controlled motion. Through FFT-based modeling, Frame2Freq consistently reallocates discriminative capacity into the frequency regions where action-specific signals are strongest, enabling far more reliable recognition of fine-grained and nearly symmetric actions than existing temporal adapters.

\section{Experiments}
\label{sec:experiments}
\subsection{Evaluation Setup}
We evaluate our method on five datasets: SSv2\cite{ssv2}, Diving48\cite{diving48}, Drive\&Act \cite{drive_and_act}, IKEA-ASM \cite{IKEA_ASM}, and HRI-30 \cite{hri_30} spanning both large-scale fine-grained datasets and smaller, domain-specific scenarios.   All experiments use frozen CLIP \cite{CLIP} and DINOv2 \cite{dinov2} backbones (ViT-B/16, ViT-L/14), training only the adapter and linear head. We uniformly sample 16 or 32 frames for SSv2 and Diving48, and 16 for the domain-specific datasets, following official evaluation and symmetric action splits from~\cite{STEP}. 
 For few-shot settings, we adopt the SSv2-Small and SSv2-Full protocols with 1-shot and 5-shot subsets per class. Further implementation details are provided in the supplementary.

\subsection{Results for Fine  Body Movement Analysis}

\noindent\textbf{Dataset.}
We use Diving48~\cite{diving48} as our testbed for fine-grained analysis of body movement.
It contains $>18K$ segmented dive clips from major diving competitions. 
The 48 classes are defined by subtle combinations of take-off type, number of somersaults, twists, and body positions, making the dataset  challenging and requiring precise modeling of motion dynamics at different frequency bands.

\noindent\textbf{Results.}
Table~\ref{table:diving48_eval} reports results on Diving-48~\cite{diving48}.
We compare Frame2Freq against 1) our main PEFT baseline ST-Adapter \cite{St_adaptor}, on top of which we build 2) other parameter-efficient fine-tuning approaches such as AIM \cite{aim} and DualPath \cite{dualpath} and 3) fully fine-tuned framework, which we include for completeness but which are not directly comparable as they update far more parameters.
Frame2Freq-MS achieves 92.2\% top-1 accuracy, outperforming ST-Adapter by +1.8\%. 
Relative to other PEFT methods like AIM and DualPath, Frame2Freq-MS yields gains of roughly +3.5\%  while using a similar or smaller number of trainable parameters.
Although fully fine-tuned models such as ORViT reach 88.0\% top-1 accuracy with substantially more trainable parameters and an additional tracking model, Frame2Freq-MS still surpasses them by about +4-5\% while training less than 10\%  parameters.
We observe that Frame2Freq-MS is significantly stronger than our single-scale variant Frame2Freq-ST, confirming the benefit of multi-scale frequency adaptation for fine-grained motion.
Nevertheless, Frame2Freq-ST remains attractive in scenarios dominated by a single characteristic temporal scale and requires fewer trainable parameters.
Our experiments on smaller fine-grained datasets with single-scale actions further indicate that Frame2Freq-ST is competitive and sometimes stronger than Frame2Freq-MS in such settings  (see Supplementary material for detailed variant selection guidance).
Overall, these results demonstrate that adapting VFMs in frequency space is beneficial for fine-grained motion understanding.

\begin{table}[t!]
\centering
 \fontsize{8}{10}\selectfont
\setlength{\tabcolsep}{5pt}
\begin{tabular}{l c cccccc}

\toprule

\textbf{Method} & \textbf{Backbone}&\textbf{Pretrain}& \textbf{Param } 
 & \textbf{Top-1}\\

&&&\textbf{(M)} &\textbf{Acc.} 
  \\

\midrule
\addlinespace[2pt]
\multicolumn{6}{l}{\cellcolor{gray!20}\textbf{{Fully fine-tuned Frameworks}}} \\
\textcolor{gray}{TimeSformer}\cite{timesformer} 
  & \textcolor{gray}{ViT-L/14} 
  & \textcolor{gray}{IN-21K} 
  & \textcolor{gray}{121}  
  & \textcolor{gray}{81.0} \\
\textcolor{gray}{VideoSwin}\cite{liu2022video_swin} 
  & \textcolor{gray}{ViT-B/16} 
  & \textcolor{gray}{IN-21K}  
  & \textcolor{gray}{88} 
  & \textcolor{gray}{81.9} \\
\textcolor{gray}{BEVT}\cite{bevt} 
  & \textcolor{gray}{ViT-B/16} 
  & \textcolor{gray}{K400}
  & \textcolor{gray}{88} 
  & \textcolor{gray}{86.7}  \\
\textcolor{gray}{SIFAR} \cite{sifar}
  & \textcolor{gray}{ViT-B/14} 
  & \textcolor{gray}{IN-21K}
  & \textcolor{gray}{87} 
  & \textcolor{gray}{87.3}  \\
\textcolor{gray}{ORViT} \cite{OR_vit} 
  & \textcolor{gray}{ViT-B/16} 
  & \textcolor{gray}{IN-21K} 
  & \textcolor{gray}{160} 
  & \textcolor{gray}{88.0}\\ 

\multicolumn{6}{l}{\cellcolor{gray!20}\textbf{{Parameter Efficient Fine-tuning Frameworks}}} \\

AIM \cite{aim}&ViT-B/16 & CLIP  &  11 & 88.9\\
DualPath\cite{dualpath} &ViT-B/16 & CLIP   & 10 & 88.7  \\
ST-Adapter \cite{St_adaptor}&ViT-B/16 & CLIP  &  7 & 90.4\\
\cdashline{1-8}[2pt/3pt] 
\addlinespace[2pt]
\textbf{Frame2Freq-ST (Ours)}  &ViT-B/16 & CLIP   & 3.6 & 75.1 \\
\textbf{Frame2Freq-MS (Ours)}  &ViT-B/16 & CLIP   &7.3 & \textbf{92.2 }\\

\bottomrule
\end{tabular}
\caption{Results on Diving48~\cite{diving48} testbed (classifying fine-grained stages of   diving actions), setting new state-of-the-art.
Here we observe a largest gain  from multi-scale modelling (Frame2Freq-MS), likely stemming from the  complex multi-component body movements  of this benchmark. 
Frame2Freq-ST, despite lower accuracy, remains interesting for resource-constrained settings due to requiring only one-third of the trainable parameters.
}

\vspace{-1em}
\label{table:diving48_eval}
\end{table}

\subsection{Results for Fine-grained Human-Object Interaction Recognition}
\noindent\textbf{Datasets.}
 Next, we evaluate our method on three action recognition datasets comprising a high amount of fine-grained human-object interactions from diverse domains:   driver monitoring (Drive\&Act \cite{drive_and_act}),  assembly tasks (IKEA-ASM \cite{IKEA_ASM}), and human-robot interaction  (HRI-30 \cite{hri_30}).
 One characteristic of these datasets is the large number of nearly symmetric action pairs~\cite{STEP} that are difficult to distinguish, such as \textit{opening} vs. \textit{closing bottle} and \textit{pick up leg} vs. \textit{lay down leg}.
 Since such actions differ only in very subtle temporal patterns, these datasets provide an excellent application scenario for our frequency-aware adapters.
 These datasets are also comparably smaller size. Drive\&Act contains $\sim$3.5k clips across 34 activities, IKEA-ASM provides $\sim$5.9k clips covering 33 assembly actions, and HRI-30 consists of  $\sim$2.9k clips across 30 fine-grained human–robot interactions. All three datasets are rather small compared to the other two datasets we consider. 

\begin{table*}[tbh]
\centering
\fontsize{9}{11}\selectfont
\setlength{\tabcolsep}{4.5pt}
\begin{tabular}{
l c c
 c c c
!{\vrule width 1pt}
c c c
!{\vrule width 1pt}
c c c
}
\toprule
\multirow{2}{*}{\textbf{Method}} & \multirow{2}{*}{\textbf{Pretrain}} & \multirow{2}{*}{\textbf{Param (M)}} &
\multicolumn{3}{c}{\textbf{Drive\&Act}} & 
\multicolumn{3}{c}{\textbf{IKEA-ASM}} & 
\multicolumn{3}{c}{\textbf{HRI-30}} \\
\cmidrule(lr){4-6} \cmidrule(lr){7-9} \cmidrule(lr){10-12}
 & & & \textbf{Sym} & \textbf{N-Sym} & \textbf{Top-1} &
 \textbf{Sym} & \textbf{N-Sym} & \textbf{Top-1} &
 \textbf{Sym} & \textbf{N-Sym} & \textbf{Top-1}\\
\midrule

\multicolumn{12}{l}{\cellcolor{gray!20}\textbf{{Fully Fine-tuned Frameworks}}}\\
\textcolor{gray}{VideoSwin-B} \cite{liu2022video_swin} 
  & \textcolor{gray}{K400} 
  & \textcolor{gray}{88} 
  & \textcolor{gray}{–} 
  & \textcolor{gray}{–} 
  & \textcolor{gray}{72.9} 
  & \textcolor{gray}{–} 
  & \textcolor{gray}{–} 
  & \textcolor{gray}{72.6} 
  & \textcolor{gray}{–} 
  & \textcolor{gray}{–} 
  & \textcolor{gray}{–}\\
\textcolor{gray}{Uniformerv2} \cite{uniformerv2} 
  & \textcolor{gray}{IN-21K} 
  & \textcolor{gray}{115} 
  & \textcolor{gray}{–} 
  & \textcolor{gray}{–} 
  & \textcolor{gray}{76.7} 
  & \textcolor{gray}{–} 
  & \textcolor{gray}{–} 
  & \textcolor{gray}{–} 
  & \textcolor{gray}{–} 
  & \textcolor{gray}{–} 
  & \textcolor{gray}{–}\\
\textcolor{gray}{SlowOnly} \cite{hri_30} 
  & \textcolor{gray}{K400} 
  & \textcolor{gray}{32} 
  & \textcolor{gray}{–} 
  & \textcolor{gray}{–} 
  & \textcolor{gray}{–} 
  & \textcolor{gray}{–} 
  & \textcolor{gray}{–} 
  & \textcolor{gray}{–} 
  & \textcolor{gray}{–} 
  & \textcolor{gray}{–} 
  & \textcolor{gray}{86.6}\\


\multicolumn{12}{l}{\cellcolor{gray!20}\textbf{{Parameter-Efficient Fine-tuning (PEFT)}}}\\
M2-CLIP \cite{m2_clip} & CLIP & 14.8 & 67.4 & 85.7 & 77.2 & 76.2 & 72.6 & 71.9 & 81.8 & 95.1 & 85.9\\
M2-CLIP \cite{m2_clip} & DINOv2 & 14.8 & 68.4 & 85.0 & 77.7 & 75.3 & 74.6 & 74.9 & 78.4 & 84.6 & 80.5\\
VitaCLIP \cite{VitaCLIP} & CLIP & 29.0 & 65.5 & 81.9 & 73.9 & 73.1 & 58.6 & 63.3 & 60.5 & 91.8 & 71.0\\
VitaCLIP \cite{VitaCLIP} & DINOv2 & 29.0 & 65.5 & 79.8 & 72.9 & 74.2 & 72.6 & 72.5 & 49.7 & 86.8 &62.0\\
ST-Adapter \cite{St_adaptor} & CLIP & 7.1 & 55.5 & \underline{88.4} & 72.6 & 70.3 & 70.7 & 70.2 & 73.8 & 95.7 & 81.1\\
ST-Adapter \cite{St_adaptor} & DINOv2 & 7.1 & 66.4 & 83.4 & 75.2 & 68.5 & 73.9 & 70.5 & 81.5 & 95.3 & 85.5\\
\cdashline{1-12}[2pt/3pt]
\addlinespace[2pt]
\textbf{Frame2Freq-ST (Ours)} & CLIP & 3.5 & 75.5 & 86.9 & 81.4 & 75.6 & 66.1 & 70.1 & 75.4 & 97.9 & 82.9\\
\textbf{Frame2Freq-ST (Ours)} & DINOv2 & 3.5 & \underline{76.9} & 86.8 & \textbf{82.0} & 80.3 & \textbf{77.0} & \textbf{78.1} & \underline{82.3} & \underline{98.6} & \underline{87.7}\\
\textbf{Frame2Freq-MS (Ours)} & CLIP & 7.3 & 64.4 & \textbf{88.6} & 77.4 & \textbf{83.9} & \underline{75.1}& \underline{78.0} & 79.1 & 98.9 & 85.7\\
\textbf{Frame2Freq-MS (Ours)} & DINOv2 & 7.3 & \textbf{77.1} & 85.5 & \underline{81.5} & \underline{81.0} & 74.2 & 76.4 & \textbf{85.6} & \textbf{98.9} & \textbf{89.8}\\

\bottomrule
\end{tabular}
\caption{Comparison across fine-grained, domain-specific action recognition datasets with common human–object interactions: Drive\&Act (driver monitoring), IKEA-ASM (furniture assembly), and HRI-30 (human–robot collaboration). Frame2Freq-ST and Frame2Freq-MS consistently surpass PEFT, and fully fine-tuned baselines while using 2-3$\times$ fewer parameters. Gains are especially strong for symmetric actions, where subtle temporal differences benefit most from frequency-based modeling.}

\label{tab:combined_results}
\end{table*}


\begin{table}[ht!]
\centering
\resizebox{\linewidth}{!}{
 \fontsize{9}{12}\selectfont
\setlength{\tabcolsep}{5.5pt}
\begin{tabular}{l c ccccc}

\toprule

\textbf{Method} 
& \textbf{\shortstack{Back-\\bone}} 
& \textbf{\shortstack{Giga\\FLOP}} 
& \textbf{\shortstack{Train.\\Par. (M)}} 
& \textbf{Views} 
& \textbf{\shortstack{Top-1\\Acc.}} 
& \textbf{\shortstack{Top-5\\Acc.}} \\

\midrule
\addlinespace[2pt]
\multicolumn{7}{l}{\cellcolor{gray!20}\textbf{{Fully fine-tuned Frameworks}}} \\
\textcolor{gray}{TimeSformer}$^\dagger$\cite{timesformer} 
  & \textcolor{gray}{ViT-L/14} 
  & \textcolor{gray}{1703}
  & \textcolor{gray}{121}
  & \textcolor{gray}{64*1*3} 
  & \textcolor{gray}{62.4} 
  & \textcolor{gray}{-} \\

\textcolor{gray}{VideoSwin}$^\dagger$\cite{liu2022video_swin} 
  & \textcolor{gray}{ViT-B/16} 
  & \textcolor{gray}{321} 
  & \textcolor{gray}{88} 
  & \textcolor{gray}{32*1*3}
  & \ \textcolor{gray}{69.6}
  & \textcolor{gray}{-}  \\

\textcolor{gray}{MTV}$^\dagger$\cite{mtv} 
  & \textcolor{gray}{ViT-B/16} 
  & \textcolor{gray}{4790} 
  & \textcolor{gray}{310} 
  & \textcolor{gray}{32*4*3}
  & \ \textcolor{gray}{67.6}
  & \textcolor{gray}{-}  \\

\textcolor{gray}{Uniformerv2} \cite{uniformerv2}
  & \textcolor{gray}{ViT-L/14} 
  & \textcolor{gray}{2600} 
  & \textcolor{gray}{574} 
  & \textcolor{gray}{16*1*3}
  & \textcolor{gray}{\textbf{72.1}}
  & \textcolor{gray}{\textbf{93.6}}  \\

\textcolor{gray}{STAN-self} \cite{STAN} 
  & \textcolor{gray}{ViT-B/16} 
  & \textcolor{gray}{1376}  
  & \textcolor{gray}{-}  
  & \textcolor{gray}{16*1*3}
  & \textcolor{gray}{69.5}
  & \textcolor{gray}{92.7}  \\

\textcolor{gray}{DTF}$^\dagger$ \cite{DTF} 
  & \textcolor{gray}{SWIN-B} 
  & \textcolor{gray}{266} 
  & \textcolor{gray}{88} 
  & \textcolor{gray}{64*1*3}
  & \textcolor{gray}{70.1}
  & \textcolor{gray}{93.2}  \\

\textcolor{gray}{ILA} \cite{ILA} 
  & \textcolor{gray}{ViT-B/16} 
  & \textcolor{gray}{438} 
  & \textcolor{gray}{-} 
  & \textcolor{gray}{16*3*4}
  & \textcolor{gray}{66.8}
  & \textcolor{gray}{90.3}  \\

\multicolumn{7}{l}{\cellcolor{gray!20}\textbf{{Parameter Efficient Fine-tuning Frameworks}}} \\
EV-L \cite{EVL} & ViT-B/16 & 1777 & 86 & 32*1*3 & 62.4 & - \\

AIM \cite{aim} & ViT-B/16 & 819 & 14 & 32*1*3 & 69.1 & 92.2 \\

Omni-CLIP\cite{omniclip}  & ViT-B/16 & 255 & 15 & 16*4*3 & 67.3 & -\\

VitaCLIP \cite{VitaCLIP} & ViT-B/16 & 1128 & 29 & 16*1*3 & 48.7 & -  \\

M2-CLIP \cite{m2_clip} & ViT-B/16 & 842 & 16 & 32*1*3 & 69.1 & 91.8 \\

DualPath\cite{dualpath} & ViT-B/16 & 716 & 13 & 32*1*3 & 70.3 & 92.9 \\
DualPath\cite{dualpath} & ViT-L/14 & 1713 & 33 & 16*1*3 & 70.2 & 92.7 \\
ST-Adapter \cite{St_adaptor} & ViT-B/16 & 651 & 14 & 32*1*3 & 69.5 & 92.6 \\

\cdashline{1-7}[2pt/3pt] 
\addlinespace[2pt]
\textbf{Frame2Freq-MS }  & ViT-B/16 & 629  & 14 & 32*1*3 & 70.4 & 92.8 \\
\textbf{Frame2Freq-MS}  & ViT-L/14 & 1322 & 19 & 16*1*3 & \textbf{72.1} & \textbf{93.2}\\

\bottomrule
\end{tabular}
}
\caption{Comparison of action recognition accuracy, compute, and trainable parameters on SSv2. All methods except for the ones marked with  $^\dagger$ use pretrained CLIP-weights.  Methods marked with $^\dagger$ use different pretraining source: TimeSformer, VideoSwin, and MTV on ImageNet-21K, and DTF-Transformer on Kinetics-400.}
\label{table:ssv2_eval}
\vspace{-1.2em}
\end{table}

\noindent\textbf{Results.}
In Table \ref{tab:combined_results}, we compare results for 1) all actions, 2) only symmetric pairs which are especially hard to distinguish, and 3) non-symmetric actions following the categorization of~\cite{STEP}. 
On Drive\&Act we surpass conventional PEFT methods by +9–11\%  for symmetric actions and +4–9\%  in overall Top-1 accuracy.
Similar trends appear on IKEA-ASM, where Frame2Freq-ST and Frame2Freq-MS deliver +3–7\% gains. 
On HRI-30, Frame2Freq continues to outperform strong PEFT baselines, delivering +4\% improvement in both symmetric and overall accuracy. 
 Across both CLIP \cite{CLIP} and DINOv2 \cite{dinov2}, improvements remain steady, with DINOv2  showing an additional +1–2\% gain on average, indicating that the spectral adaptation generalizes across distinct pretraining objectives. 
Interestingly, in this task, the considerably smaller Frame2Freq-ST variant proves highly effective: it performs on par with, and in some cases slightly better than, the larger Frame2Freq-MS.
We attribute this behavior to the dominance of a single characteristic temporal scale in these domain-specific benchmarks, which aligns well with the design of the single-scale variant.
Overall, these results show that Frame2Freq is especially effective for fine-grained, nearly symmetric actions and excels in such domain-specific scenarios.

\subsection{Results for General Human-Object Interaction Recognition}

\noindent\textbf{Dataset.}
The Something-Something v2 dataset (SSv2) ~\cite{ssv2} contains $\sim$220k crowd-sourced videos of everyday hand–object interactions with comparatively coarse action labels, and is larger than the previous domain-specific benchmarks in terms of both the number of clips and visual variability. 
On SSv2, prior work evaluates models both in the full-dataset regime and in 1-shot and 5-shot settings. Accordingly, we also adopt both evaluation regimes.

\noindent\textbf{Full Dataset Results.}
Table~\ref{table:ssv2_eval} reports results on SSv2, a large-scale benchmark of short, motion-heavy interactions with comparatively coarse action labels.
Frame2Freq-MS improves over our main baseline ST-Adapter~\cite{St_adaptor} by +0.9\% top-1 accuracy at similar computational cost and outperforms other PEFT methods by up to +3.0\%.
We note that SSv2 is the setting where our gains are smallest, which we attribute to the coarse action definitions that leave less room for fine-grained frequency modeling.
Nevertheless, Frame2Freq-MS achieves state-of-the-art performance among PEFT approaches and performs on par with fully fine-tuned models such as Uniformerv2~\cite{uniformerv2}, VideoSwin~\cite{liu2022video_swin}, and DTF-Transformer~\cite{DTF}, while using less than 5\% of their trainable parameters.


\noindent\textbf{Few-shot Results.}
Table~\ref{table:few_shot_results} demonstrates that  Frame2Freq-MS yields state-of-the-art results in the data-scarce SSv2 regime.
Frame2Freq-MS achieves consistent gains of +1.5–3.0\% over fully supervised PEFT baselines such as ST-Adapter~\cite{D2_ST} and DualPath in both 1-shot and 5-shot settings on SSv2-Small and SSv2-Full.
It also matches or surpasses specialized few-shot architectures such as TATs~\cite{tats}, Trokens~\cite{trokens}, and D2ST-Adapter~\cite{D2_ST}, which rely on explicit trajectory modeling or deformable attention for motion tracking.
Our model attains 55.7\% / 71.3\% top-1 accuracy on SSv2-Small and 66.9\% / 82.0\% on SSv2-Full (1-shot / 5-shot) using a simpler frequency-aware adapter with fewer trainable parameters.

\begin{table}[t!]
\centering
\fontsize{8}{10}\selectfont
\setlength{\tabcolsep}{3.5pt}
\begin{tabular}{l c cc cc }
\toprule
\textbf{Method} & \textbf{Pretrain} 
& \multicolumn{2}{c}{\textbf{SSv2-Small}} 
& \multicolumn{2}{c}{\textbf{SSv2-Full}}
\\
\cmidrule(lr){3-4} \cmidrule(lr){5-6} 
& & \textbf{1-shot} & \textbf{5-shot} 
  & \textbf{1-shot} & \textbf{5-shot}
  \\

\midrule
AIM \cite{aim} & CLIP & 52.8 & 67.5 & 63.7 & 79.2 \\
ST-Adapter \cite{St_adaptor} & CLIP & 53.1 & 68.0 & 64.2 & 79.5 \\
DualPath \cite{dualpath} & CLIP & 53.5 & 68.1 & 64.5 & 79.8 \\
TATs \cite{tats}& DINO & 47.9 & 64.4 & 57.7 & 74.6 \\
Trokens \cite{trokens}& DINOv2 & 53.4 & 68.9 & 61.5 & 76.7 \\
D2ST-Adapter \cite{D2_ST}& CLIP & 55.0 & 69.3 & 66.7 & 81.9 \\
\textbf{Frame2Freq-MS (Ours)} & CLIP & \textbf{55.7} & \textbf{71.3} & \textbf{66.9} & \textbf{82.0} \\

\bottomrule
\end{tabular}
\caption{Few-shot results on SSv2-Small and SSv2-Full  datasets. Frame2Freq-MS achieves state-of-the-art performance.}

\label{table:few_shot_results}
\vspace{-1.2em}
\end{table}

\subsection{Ablation Studies}
Next, we analyze the impact of  spectral cues, adapter configuration, and fusion mechanisms on the recognition quality. All SSv2 and Diving48 experiments use the CLIP backbone with the Frame2Freq-MS adapter, while Drive\&Act and IKEA-ASM use the DINOv2 backbone with Frame2Freq-ST. 
All models are trained for 60 epochs with 16 frames per clip.

\noindent\textbf{Impact of  Frequency Cues on Temporal Adaptation.} 
Table~\ref{table:combined_ablation} evaluates three core design choices in Frame2Freq: (1) whether frequency cues, temporal cues, or both are used, (2) where the adapter is placed, and (3) how the two signals are fused.  
We observe that using only temporal adaptation underperforms on symmetric actions, while using only frequency cues captures global motion but misses fine timing. 
Combining both consistently achieves the best results across all datasets. 
Together, they form a complementary representation that yields the highest overall accuracy.

\begin{table}[t]
\centering
\resizebox{1.05\linewidth}{!}{
\begin{tabular}{lcccc}
\toprule
\textbf{Method} & \multicolumn{2}{c}{\textbf{Frame2Freq-ST} }& \multicolumn{2}{c}{\textbf{Frame2Freq-MS} }\\
\cmidrule(lr){2-3} \cmidrule(lr){4-5}
 & \textbf{D\&A} & \textbf{IKEA} & \textbf{SSv2} &\textbf{Diving} \\

\midrule
\multicolumn{5}{l}{\cellcolor{gray!20}\textbf{{Effect of Frequency Cues}}}\\
Only Frequency Conv. & 81.6 & 75.6 & 67.5 &90.9 \\
Only Temporal Conv. & 75.2 & 70.1 & 69.1 & 90.4 \\
Combined (Frame2Freq) & \textbf{82.0} & \textbf{78.1} & \textbf{69.7} & \textbf{92.2} \\
\midrule
 \multicolumn{5}{l}{\cellcolor{gray!20}\textbf{{ Adapter placement}}}\\
 Before MHSA & 80.4 & 76.1 & 68.6 &91.1 \\
 After MHSA & \textbf{82.0} & \textbf{78.1} & 68.3 &\textbf{92.2} \\
 Both & 81.7 & 75.4 & \textbf{69.7} &90.8 \\
 \midrule
 \multicolumn{5}{l}{\cellcolor{gray!20}\textbf{{Fusion strategy}}}\\
 Gated Fusion & 81.4 & 76.2 & 69.1&91.7\\
 Learnable Fusion & 81.9 & 76.6 &68.9 &91.5\\
 Mean/Concat& \textbf{82.0} & \textbf{78.1} & \textbf{69.7} &\textbf{92.2} \\
  \multicolumn{5}{l}{\cellcolor{gray!20}\textbf{{ Multi-scale window size}}}\\
$[T]$ &\texttt{N/A}&\texttt{N/A}& 69.0 & 91.5 \\
$[T, T/2]$ &\texttt{N/A}&\texttt{N/A}&  69.3 & 91.6 \\
$[T, T/2, T/4]$ &\texttt{N/A}&\texttt{N/A}&  \textbf{69.7} & \textbf{92.2} \\
$[T, T/2, T/4, T/8]$ &\texttt{N/A}&\texttt{N/A}&  69.4 & 91.0 \\
 \bottomrule
\end{tabular}
}
\caption{Ablation of different design choices. Combining frequency and temporal modules yields the best results, confirming their complementary strengths. Placing adapters \textit{after} MHSA performs strongest. For  fusion strategy, simple mean/concat fusion is most effective. A three–scale setup $[T,\; T/2,\; T/4]$  performs best, balancing temporal detail with broader context. Multi-scale window ablations are \texttt{N/A} for ST variant.}
\label{table:combined_ablation}
\vspace{-1.2em}
\end{table}

\noindent\textbf{Impact of Adapter Relation to MHSA.}
Table \ref{table:combined_ablation} (second experimental group) examines how adapter position affects representation learning within the transformer block. Post-MHSA insertion delivers the highest gains on Drive\&Act, IKEA-ASM, and Diving48 (+1-2\%), as it refines temporally aggregated features after spatial attention.  SSv2 benefits from dual placement before and after MHSA, since its motion-dense clips require deeper temporal integration. 


\noindent\textbf{Impact of Fusion Strategy.}
Table~\ref{table:combined_ablation} evaluates how different mechanisms combine frequency and temporal cues. Simple mean/concatenation outperforms gated and learnable fusion. This suggests that the two cues are already highly complementary, and that heavier fusion modules may over-parameterize the interaction rather than enhancing it.

\noindent\textbf{Impact of Adapter location.}
Table \ref{tab:placement_ablation_combined} examines adapter placement across transformer depth. Performance improves as adapters are introduced in higher layers, with the best results obtained when distributed throughout all stages. Restricting adapters to early layers leads to a sharp drop due to limited semantic abstraction, while excluding them entirely from early layers also degrades performance. This shows that shallow layers also contribute essential motion cues, and balancing early-layer motion cues with deep-layer semantic refinement is the most effective for fine-grained video understanding.


\noindent\textbf{Effect of Multi-scale Window Size.} 
Table~\ref{table:combined_ablation} analyzes the influence of multi-scale temporal windows on performance. 
Expanding from a single scale $[T]$ to three scales $[T, T/2, T/4]$ progressively improves results, as it captures slow, mid, and fast motion bands and thus enriches the temporal representation. 
However, adding an even finer scale $T/8$ no longer yields improvements, indicating that the benefit of additional scales saturates once the relevant motion frequencies are already covered.

\begin{table}[t]
\centering
\renewcommand{\arraystretch}{1.2}
\setlength{\tabcolsep}{8pt}
\resizebox{\linewidth}{!}{
\begin{tabular}{ccccccc}
\toprule
& & & \multicolumn{2}{c}{\textbf{Frame2Freq-ST} }& \multicolumn{2}{c}{\textbf{Frame2Freq-MS} }\\
 \cmidrule(lr){4-5} \cmidrule(lr){6-7}
 \textbf{1--4}& \textbf{5--8} & \textbf{9--12} & \textbf{D\&A} & \textbf{IKEA} & \textbf{SSv2} &\textbf{Diving48} \\
\midrule
\checkmark &            &            & 49.9 & 57.6 & 55.8 & 67.6 \\
           & \checkmark &            & 77.7& 72.9 & 66.0 & 80.9 \\
           &            & \checkmark & 80.4 & 76.1 & 67.3 & 87.6 \\
\checkmark & \checkmark &            & 78.2 & 70.5 & 67.0 & 90.6 \\
           & \checkmark & \checkmark & 81.4 & 77.2 & 69.2 & 91.2 \\
\checkmark & \checkmark & \checkmark & \textbf{82.1} & \textbf{78.1} & \textbf{69.7} & \textbf{92.2} \\
\bottomrule
\end{tabular}
}
\caption{Frame2Freq-adapter placement at different network depths. Single-location placement consistently yields better results in the deeper layers (9--12). Combining placements at all three locations leads to the best outcome on all benchmarks.
}

\label{tab:placement_ablation_combined}
\vspace{-1em}
\end{table}



\section{Conclusion}
\label{sec:conclusion}

We introduced Frame2Freq, a frequency–aware adapter that enhances temporal reasoning in Vision Foundation Models through operations based on spectral frequency analysis. By performing PEFT  in the time frequency domain, our method achieves consistent gains across fine-grained and domain-specific benchmarks, setting new state-of-the-art on four fine-grained datasets and performing on par with fully fine-tuned models on the larger, more general SSv2 dataset, while still outperforming all PEFT methods across every benchmark. Our extensive ablations confirm that these improvements stem from the complementary strengths of spectral and temporal adaptation mechanisms.
 Beyond Fourier analysis, future extensions may incorporate wavelets, multiresolution filters, or learnable time-frequency operators.

\noindent \textbf{Acknowledgments.} The research published in this article is supported by the Deutsche Forschungsgemeinschaft (DFG) under Germany’s Excellence Strategy – EXC 2120/1 –390831618. The authors also thank the International Max Planck Research School for Intelligent Systems (IMPRS-IS) for supporting Thinesh Thiyakesan Ponbagavathi.
The authors also gratefully acknowledge the computing time provided on the high-performance computer HoreKa by the National High-Performance Computing Center at KIT. 
This center is jointly supported by the Federal Ministry of Education and Research and the Ministry of Science, Research and the Arts of Baden-Württemberg, as part of the National High-Performance Computing (NHR) joint funding program (https://www.nhr-verein.de/en/our-partners). 
HoreKa is partly funded by the German Research Foundation (DFG). 

{
  \small
\bibliographystyle{ieeenat_fullname}
 \bibliography{main}
}

 \clearpage
\setcounter{page}{1}
\renewcommand{\thesubsection}{\Alph{subsection}}
\maketitlesupplementary
\subsection{Implementation Details}
\subsubsection{Frequency Spectral Analysis} 
\noindent\textbf{Video Spectral Analysis.} To visualize and compare the motion characteristics of different action classes, we compute full 3D Fourier spectra directly from raw videos. Each video is first uniformly sampled to 16 grayscale frames and resized to 224×224 pixels. This produces a spatiotemporal volume $V \in \mathbb{R}^{H \times W \times T}$ with H=W=224 and T=16. A 3D FFT is then applied to the entire tensor,
$F = \text{FFT}_{(x,y,t)}(V)$, followed by a centered frequency shift. The magnitude (or power) spectrum is log-compressed to stabilize large dynamic ranges. To emphasize discriminative patterns, we optionally remove the DC component and use spectral whitening (i.e., division by the radial frequency radius). We take the central temporal slice of the 3D spectrum, which gives a stable 2D frequency map that captures spatial–temporal motion patterns without mixing across temporal bands. Finally, we calculate the class-wise mean spectra to verify if the patterns stay relevant across multiple samples in the dataset. We perform this by averaging FFT magnitudes across all videos in a class after normalizing their shapes and exporting them as high-resolution maps.
\begin{algorithm}[ht!]
\caption{Frequency Discriminability Analysis}
\label{alg:freq_disc}
\KwIn{Temporal embeddings $\{X_i \in \mathbb{R}^{T \times D}\}_{i=1}^N$, class labels $\{y_i\}_{i=1}^N$}
\KwOut{Normalized discriminability scores $\{\hat{D}(f)\}_{f=1}^F$}

\BlankLine
\textbf{Step 1: Compute spectral power per clip} \\
\ForEach{clip $i$}{
    $\tilde{X}_i(f) \leftarrow \text{FFT}_t(X_i)$ \tcp*{1D FFT along temporal axis}
    $P_i(f) \leftarrow \mathbb{E}_d \, |\tilde{X}_i(f,d)|^2$ \tcp*{Spectral power (mean over $D$)}
}
\BlankLine
\textbf{Step 2: Group by class labels} \\
Let $\mathcal{C}$ be the set of unique classes. \\
For each $c \in \mathcal{C}$, define $\mathcal{P}_c(f) = \{ P_i(f) \mid y_i = c \}$.
\BlankLine
\textbf{Step 3: Compute per-frequency discriminability (ANOVA principle)} \\
\For{$f = 1$ \KwTo $F$}{
    $\mu(f) \leftarrow \frac{1}{N} \sum_i P_i(f)$ \tcp*{Global mean}
    \For{$c \in \mathcal{C}$}{
        $\mu_c(f) \leftarrow \frac{1}{|\mathcal{P}_c|} \sum_{p \in \mathcal{P}_c(f)} p$ \tcp*{Class-wise mean}
    }
    $\text{Between}(f) \leftarrow \sum_{c \in \mathcal{C}} |\mathcal{P}_c| \, [\mu_c(f) - \mu(f)]^2$ \\
    $\text{Within}(f) \leftarrow \sum_{c \in \mathcal{C}} \sum_{p \in \mathcal{P}_c(f)} [p - \mu_c(f)]^2$ \\
    $D(f) \leftarrow \frac{\text{Between}(f)}{\text{Within}(f) + \epsilon}$ \tcp*{Frequency-wise discriminability}
}
\BlankLine
\textbf{Step 4: Normalize across frequencies} \\
$\hat{D}(f) \leftarrow \frac{D(f)}{\sum_{f'=1}^{F} D(f')}$ \tcp*{Normalize such that $\sum_f \hat{D}(f) = 1$}
\BlankLine
\Return $\{\hat{D}(f)\}_{f=1}^F$
\label{algo:freq_discrim}
\end{algorithm}

\noindent\textbf{Frequency Discriminability Analysis.} To quantify how informative each frequency is for distinguishing actions, we perform a Frequency Discriminability Analysis inspired by Analysis of Variance (ANOVA) \cite{annova}. Temporal embeddings $X_i \in \mathbb{R}^{T \times D}$ extracted from the backbone (after adapter insertion) are transformed using a 1D FFT along time, giving $\tilde{X}i(f)$. This operation decomposes each embedding dimension into frequency components, capturing periodicity, phase changes, and motion rhythms encoded by the model. We convert these into scalar frequency responses by computing the mean spectral power across embedding dimensions,
$P_i(f) = \mathbb{E}_d |\tilde{X}_i(f, d)|^2$, resulting in a power vector for each clip that summarizes its activity across frequencies.
Clips are then grouped according to their ground-truth class labels, allowing us to compute class-wise means and within-class variances at each frequency. Following the ANOVA principle, the discriminability score is obtained by taking the ratio between the between-class and within-class variance at each temporal frequency,
$D(f) = \frac{\text{Between}(f)}{\text{Within}(f) + \epsilon}$,where a small $\epsilon$ ensures numerical stability. Finally, the discriminability curve is normalized across frequencies to produce $\hat{D}(f)$, which represents a probability-like distribution indicating the amount of discriminative power present at each temporal band.
Algorithm \ref{algo:freq_discrim} provides an overview of the implementation of our Frequency Discriminability Analysis.

\subsubsection{Model Implementation}
 All our models are trained using frozen CLIP \cite{CLIP} or DINOv2 \cite{dinov2} backbones, updating only the Frame2Freq adapter and the linear classification head. Training is performed on 16x NVIDIA A100 GPUs for SSv2 \cite{ssv2} and Diving48 \cite{diving48}, whereas 4x NVIDIA A100 GPUs where used for training the fine-grained human-object interaction datasets ( Drive\&Act \cite{drive_and_act}, IKEA-ASM \cite{IKEA_ASM}, and HRI-30 \cite{hri_30}) with DDP used for GPU-parallelization. We train for 60 epochs with AdamW, cosine decay, batch size 32, learning rate 1e-3, weight decay 0.05, and spatial resolution 224×224. We also use 2 warmup epochs for all the datasets to ensure smooth learning of parameters. FFT operations are computed using batched CUDA kernels without additional regularization or test-time augmentation. This setup ensures that all reported improvements arise solely from the frequency-aware adapter and its integration into the frozen spatial backbone, isolating the contribution of spectral modeling to fine-grained temporal understanding. 
 
\noindent\textbf{Implementation of Frequency Transforms.} 
We apply STFT per spatial location and channel by reshaping features to $(B \cdot H \cdot W \cdot C_a, T)$ and using a Hann window with $n_{fft} = min(32, T/2)$ and $hop = n_{fft}/4$ ($\sim$ 75\% overlap) and returns complex tensors. Complex spectra are processed by concatenating real/imaginary parts (phase preserved), refined via depthwise Conv3D in the time–frequency grid, and reconstructed with iSTFT using identical parameters.

We now describe the dataset-specific preprocessing and sampling strategies used across all benchmarks.

\noindent\textbf{Diving48.}
Diving48 \cite{diving48} contains fast, tightly synchronized diving sequences where frame-to-frame changes are minimal, so fine-grained temporal cues are crucial. We uniformly sample 32 frames per clip and apply RandAugment with random resized cropping and horizontal flips. Frame2Freq is inserted after the MHSA layer in every transformer block. At inference, we use a single temporal view with three spatial crops, matching standard PEFT evaluation practice. 

\noindent\textbf{Something Somethingv2.} Something-Something V2 \cite{ssv2} is built around short, object-centric interactions where motion speed varies sharply across classes. We uniformly sample 16 or 32 frames and apply the standard SSv2 augmentations: RandAugment, random resized crops, and color jitter. Following prior PEFT setups, we use two Frame2Freq adapters per block for ViT-B/16 and a single adapter for ViT-L/14. In inference, we use one temporal clip and three spatial crops to match established evaluation practice.

\noindent\textbf{Drive\&Act.} Drive\&Act \cite{drive_and_act} features in-cabin driver monitoring with fine-grained, nearly symmetric actions such as reaching, picking, or placing objects. We follow the standard Drive\&Act preprocessing pipeline: sampling 16 frames with a frame interval of 2, and applying a sequence of augmentations, including resizing, Random Resized Crop, and horizontal flips.  We insert a single Frame2Freq adapter per transformer block, as this lightweight variant avoids overfitting on the smaller domain. During inference, we use three temporal clips and a spatial center crop to maintain consistency with prior work.

\noindent\textbf{IKEA-ASM.}IKEA-ASM \cite{IKEA_ASM} contains long-horizon assembly sequences with frequent nearly symmetric motions, such as pick up vs.\ lay down components or tightening vs.\ loosening screws. We sample 16 frames per clip at a frame interval of 2 and apply the standard augmentation pipeline used in prior assembly-action work, which includes resizing, Random Resized Crop, and flips. Similar to Drive\&Act, we use one adapter per block and utilize three temporal clips per video during inference.

\noindent\textbf{HRI-30.}HRI-30 \cite{hri_30} captures close-range human–robot collaborative tasks with subtle object exchanges, handovers, and nearly symmetric motion phases that demand precise temporal cues. We uniformly sample 16 frames per clip and follow the same preprocessing protocol as Drive\&Act and IKEA-ASM:  RandomResizedCrop, horizontal flips, and per-frame normalization. Given the dataset’s small size and fine-grained temporal structure, we again employ one adapter per block and one temporal clip and three spatial crops to match established evaluation practice.

\subsection{Additional Experiments and Analysis}
\subsubsection{Variant Selection Guide}
To better understand the behavior of the two variants, we analyze performance across action categories with different temporal characteristics (Table~\ref{table:Scale_analysis}). 
On Diving48, actions involving a single characteristic temporal scale (e.g., no somersaults or twists) show comparable performance between ST and MS variants. In contrast, actions composed of multiple somersaults or twists exhibit \textit{overlapping temporal frequencies}, where Frame2Freq-MS provides clear gains. 
A similar pattern emerges on HRI-30. Frame2Freq-MS improves performance on Movement+Manipulation classes that involve multi-scale temporal dynamics (simultaneous body and hand motion), while Frame2Freq-ST performs competitively on Pure Movement and Pure Manipulation categories characterized by a dominant temporal scale. 
Overall, these findings indicate that the single-scale variant is well suited for datasets dominated by a single characteristic temporal frequency, whereas the multi-scale variant is advantageous for complex actions with overlapping temporal structures.

\begin{table}[ht!]
\centering
\scriptsize
\setlength{\tabcolsep}{3pt}

\begin{tabular}{l c c c c c c}
\toprule
 & \multicolumn{3}{c}{\textbf{HRI-30}} & \multicolumn{3}{c}{\textbf{Diving48}} \\
\cmidrule(lr){2-4} \cmidrule(lr){5-7}
\textbf{Method} 
& \textbf{Pure} & \textbf{Pure} & \textbf{Movement+}
& \textbf{Single} & \textbf{Multiple} & \textbf{Other} \\
& \textbf{Movement} & \textbf{Manip.} & \textbf{Manip.}
& \textbf{Scale} & \textbf{Scale} & \textbf{Classes} \\
\midrule
F2F-ST     & 97.9 & 95.4 & 69.1 &89.9 & 65.9 & 73.9 \\
F2F-MS     & 98.6 & 97.5 & 73.8 & 94.9 & 85.5 & 92.8 \\
\bottomrule
\end{tabular}
\caption{Performance breakdown by temporal complexity across HRI-30 and Diving48, illustrating that Frame2Freq-ST excels in single-scale settings while MS variant benefits multi-scale actions.}
\label{table:Scale_analysis}
\vspace{-1em}
\end{table}

\subsubsection{Throughput Analysis}

We compare the computational overhead of Frame2Freq-MS against existing Image-to-Video PEFT methods in Table~\ref{table:throughput}. Despite integrating frequency transforms in adapters across all transformer layers, Frame2Freq-MS maintains competitive efficiency with 7.3M trainable parameters and 314 GFLOPs. The measured inference latency (13.11 ms) is comparable to prior PEFT approaches such as ST-Adapter (12.00 ms) and remains significantly more efficient than heavier designs like VitaCLIP and M2-CLIP. These results confirm that the spectral adaptation introduces minimal runtime overhead while preserving strong empirical gains.

\begin{table}[h!]
\centering
\resizebox{0.85\linewidth}{!}{
\begin{tabular}{l c c c}
\toprule
\textbf{Method} &\textbf{Params}& \textbf{GFLOPs} & \textbf{Time (ms)} \\
\midrule
VitaCLIP \cite{VitaCLIP} & 29.0    & 1128 & 14.09 \\
M2-CLIP  \cite{VitaCLIP}  &  14.8  & 421  & 13.62 \\
ST-Adapter \cite{St_adaptor}& 7.1   & 325  & 12.00 \\
Frame2Freq-MS (Ours)       &7.3 & 314  & 13.11 \\
\bottomrule
\end{tabular}

}
\caption{Throughput Analysis of Frame2Freq in comparison to existing Image-to-Video PEFT methods.}
\label{table:throughput}
\vspace{-1em}
\end{table}

\begin{table}[ht!]
\centering
\resizebox{0.85\linewidth}{!}{
\begin{tabular}{l cccc}
\toprule
\textbf{Width} & \textbf{D\&A} & \textbf{IKEA} & \textbf{SSv2} & \textbf{Diving48} \\
\midrule
96             & 81.27 & 74.26 &   --  &   --   \\
192    & \textbf{82.04} & \textbf{78.06} & 68.9 & 91.2 \\
384     & 79.63 & 75.87 & \textbf{69.7} & \textbf{92.2} \\
768            &   --  &   --  & 69.5 & 91.1 \\
\bottomrule
\end{tabular}

}
\caption{Impact of adapter width on performance of Frame2Freq adapter in 4 datasets.}

\label{table:adapter_width}
\end{table}

\subsubsection{Impact of Adapter Width}
Table~\ref{table:adapter_width} reports the effect of varying the adapter bottleneck width across four datasets. A clear pattern emerges: smaller widths (192) perform best on fine-grained human–object datasets such as Drive\&Act and IKEA-ASM, where motion signals are subtle and larger adapters tend to overfit. In contrast, higher-capacity adapters benefit motion-rich datasets like SSv2 and Diving48, where the 384-width variant captures more complex temporal variations and achieves the highest accuracy. These results highlight that frequency-sensitive temporal adaptation does not require uniformly large adapters; instead, optimal width depends on the underlying motion complexity and size of each dataset.

\subsubsection{Frequency Analysis in Nearly Symmetric Actions}
\begin{figure}[ht!]
    \centering
\includegraphics[width=1\columnwidth]{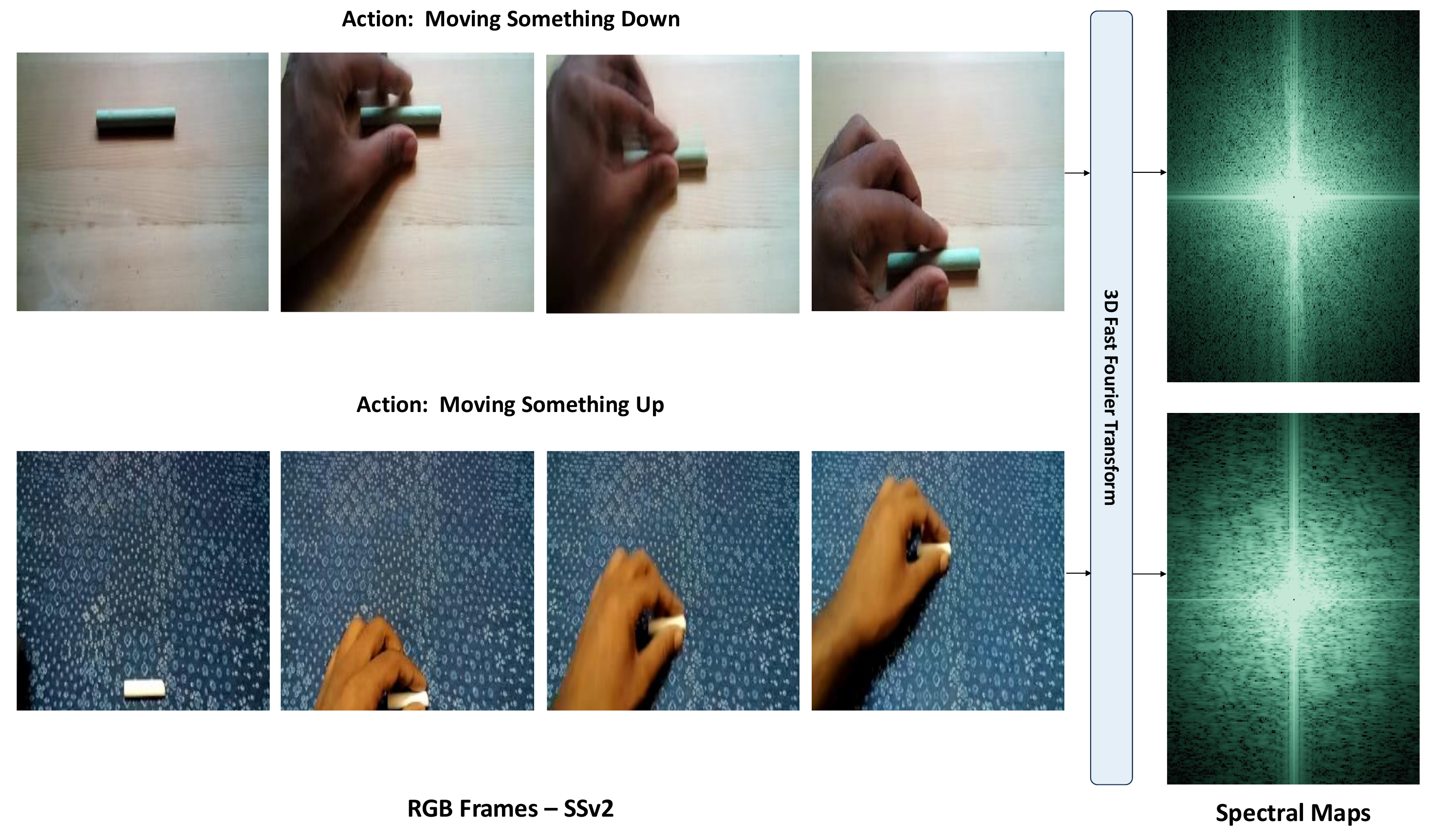}
\caption{
\textbf{SSv2 symmetric actions.} Spectral maps (right) reveal clear directional frequency differences between \textit{moving something down} and \textit{moving something up}, despite nearly identical RGB frames.
}
\label{fig:freq_ssv2}
\vspace{-1em}
\end{figure}

\begin{figure}[ht!]
    \centering
\includegraphics[width=1\columnwidth]{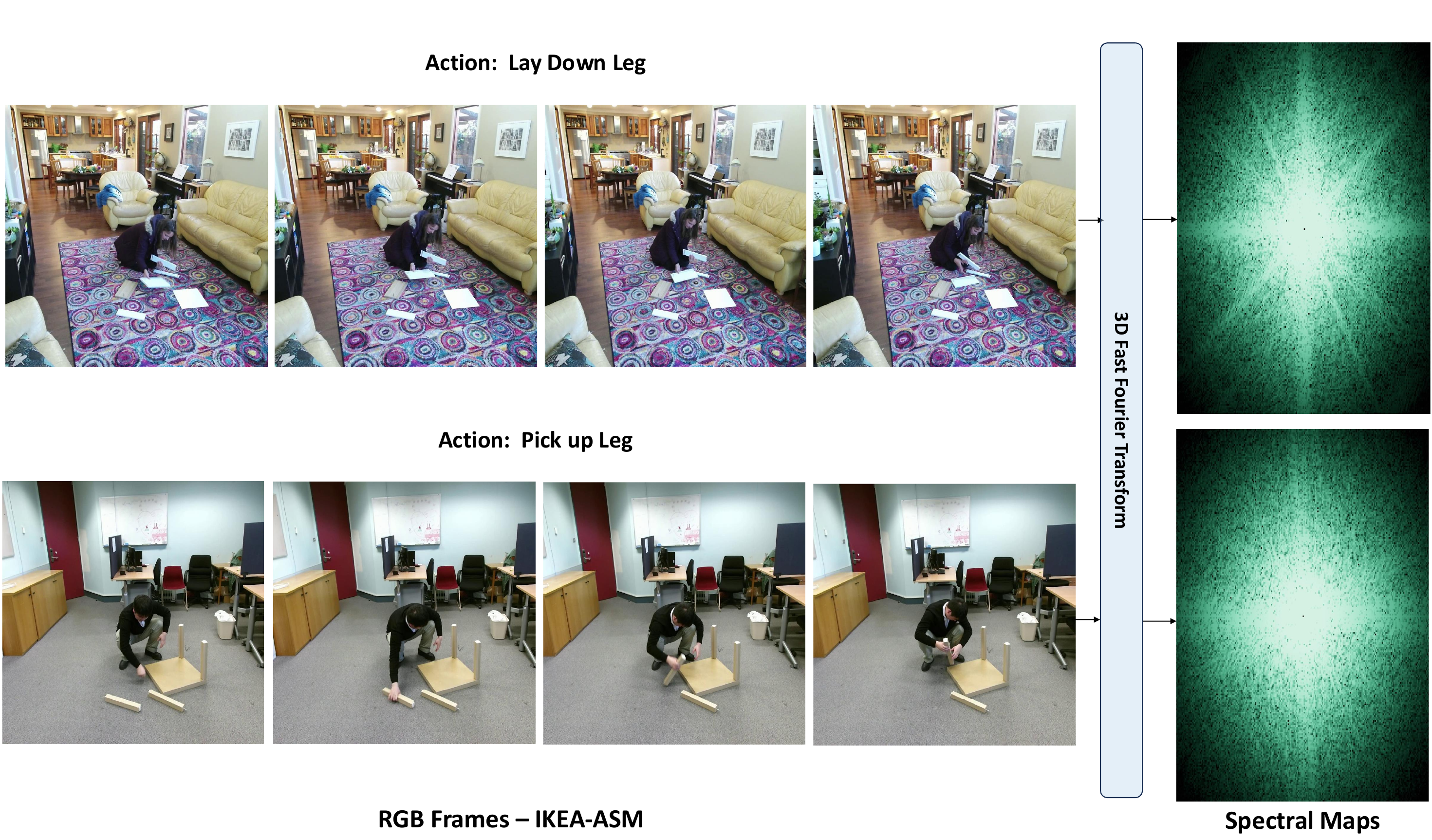}
\caption{
\textbf{IKEA-ASM symmetric actions.} 3D FFT spectra (right) distinguish \textit{lay down leg} from \textit{pick up leg} through subtle motion-direction cues that are hard to see in RGB space.
}
\label{fig:freq_ikea}
\end{figure}

\begin{figure}[ht!]
    \centering
\includegraphics[width=1\columnwidth]{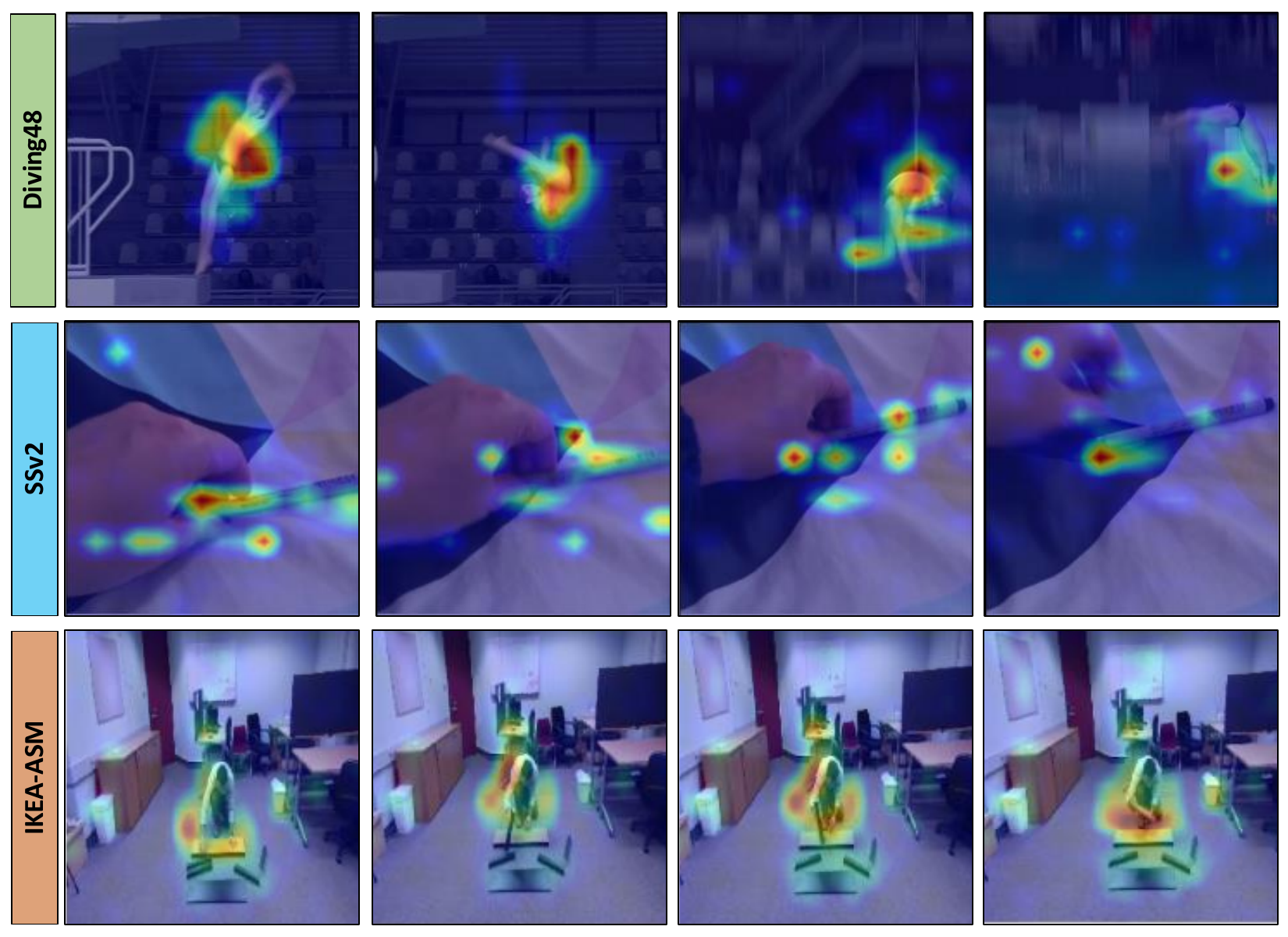}
\caption{
Cross-dataset attention visualizations. Frame-wise attention maps for Diving48, SSv2, and IKEA-ASM show that Frame2Freq consistently localizes motion-critical regions effectively.
}
\label{fig:quali}
\end{figure}
The examples from SSv2 (Figure \ref{fig:freq_ssv2}) highlight why fine-grained, direction-sensitive actions are so challenging in RGB space. Moving something up and moving something down contain nearly identical hand–object configurations and frame-to-frame appearance. Yet their 3D FFT spectra differ sharply: the frequency maps show clear orientation-specific energy shifts that mirror the direction of motion. These patterns are invisible in raw frames but emerge cleanly in the spectral domain, revealing why frequency cues can reliably disambiguate actions that defeat standard spatial reasoning.

The IKEA-ASM examples (Figure \ref{fig:freq_ikea} show the same phenomenon in a more complex human–object environment. The actions lay down leg and pick up leg share almost the same spatial layouts and pose transitions, making them nearly symmetric in RGB frames. Their spectral signatures, however, diverge: periodicity and directionality encoded in the 3D FFT expose motion structure that the eye struggles to perceive. Together, the two figures illustrate why frequency-domain representations are uniquely suited for distinguishing fine-grained, nearly mirrored actions.

\subsubsection{Qualitative Analysis}
Figure \ref{fig:quali} illustrates how Frame2Freq adapts its focus across three very different fine-grained activity datasets. In Diving48, the model concentrates on the diver’s torso and limb trajectories, capturing rotation phases that define somersault classes. In SSv2, attention locks onto the fingertips and the manipulated object, revealing the subtle contact dynamics that distinguish nearly identical hand–object motions. In IKEA-ASM, the maps consistently highlight the worker’s hands, tools, and assembly components, the regions that drive temporal progression in furniture-assembly actions. Across all settings, the heatmaps show that Frame2Freq reliably isolates the motion-critical regions, demonstrating robust cross-domain temporal grounding.

\end{document}